%% file: main.tex
\documentclass[10pt, a4paper]{article}
\usepackage{graphicx,amsfonts,booktabs, placeins,float, caption,subcaption, dirtytalk,hyperref}
\usepackage{natbib}

\title{Translation from the Information Bottleneck Perspective: an Efficiency Analysis of Spatial Prepositions in Bitexts}

\author{Antoine Taroni, Ludovic Moncla, Frédérique Laforest \
INSA Lyon, CNRS, Lyon 1 Universit\'e, LIRIS, UMR5205,\
69621 Villeurbanne, France \
fist.last@insa-lyon.fr
}

\begin{document}
\maketitle

\begin{abstract}
Efficient communication requires balancing informativity and simplicity when encoding meanings. The Information Bottleneck (IB) framework captures this trade-off formally, predicting that natural language systems cluster near an optimal accuracy-complexity frontier. 
While supported in visual domains such as colour and motion, linguistic stimuli such as words in sentential context remain unexplored.
We address this gap by framing translation as an IB optimisation problem, treating source sentences as stimuli and target sentences as compressed meanings.
This allows IB analyses to be performed directly on bitexts rather than controlled naming experiments.
We applied this to spatial prepositions across English, German and Serbian translations of a French novel.
To estimate informativity, we conducted a pile-sorting pilot-study ($N=35$) and obtained similarity judgements of pairs of prepositions. We trained a low-rank projection model ($D=5$) that predicts these judgements (Spearman $\rho=0.78$). Attested translations of prepositions lie closer to the IB optimal frontier than counterfactual alternatives, offering preliminary evidence that human translators exhibit communicative efficiency pressure in the spatial domain. More broadly, this work suggests that translation can serve as a window into the cognitive efficiency pressures shaping cross-linguistic semantic systems.
\end{abstract}


\section{Introduction}

Languages partition semantic domains in remarkably diverse ways. For example, \citet{levinson2003space} and \citet{levinson2003natural} intensively document how spatial relations are named differently by speakers of typologically diverse languages. Yet beneath this diversity, systematic regularities emerge when naming spatial relations, but also within the semantic domains of colour and kinship \textit{inter alia} \citep{review}.   
One influential standpoint invokes the principle of efficient communication to explain these regularities \citep{gibson2019efficiency, review, colours}. According to this view, when speakers of the same language balance the informational and cognitive load in their utterances, the mapping from meanings to words is optimised with respect to that trade-off between informativity and simplicity. Languages around the world achieve different realisations of this trade-off through evolution \citep{kirby2015compression}.

The Information Bottleneck (IB) framework \citep{tishby2000} was put forward by \citet{colours} to operationalise this trade-off for natural languages. It states that languages achieve near-optimal compression of meanings from an information-theoretic perspective, and has recently gained substantial empirical support. Usually, an IB analysis focuses on a specific semantic domain, and relies on naming data (e.g. natural descriptions of stimuli such as pictures or videos). The underlying assumption is that a stimulus evokes a mental representation, which is efficiently compressed into a description. The communicative cost of the description (complexity) is then compared to the relevant information it conveys about the stimuli (informativity), across natural and hypothetical languages. Prior IB applications relied on various stimuli like colour chips \citep{colours}, static visual referents from images \citep{gualdoni2023}, and kinematic events in videos of human locomotion \citep{gaits}. However, to our knowledge, no IB analysis has considered linguistic stimuli.

\begin{figure*}[!ht]
\begin{center}
\includegraphics[width=.9\textwidth, trim=  0 0 0 0, clip]{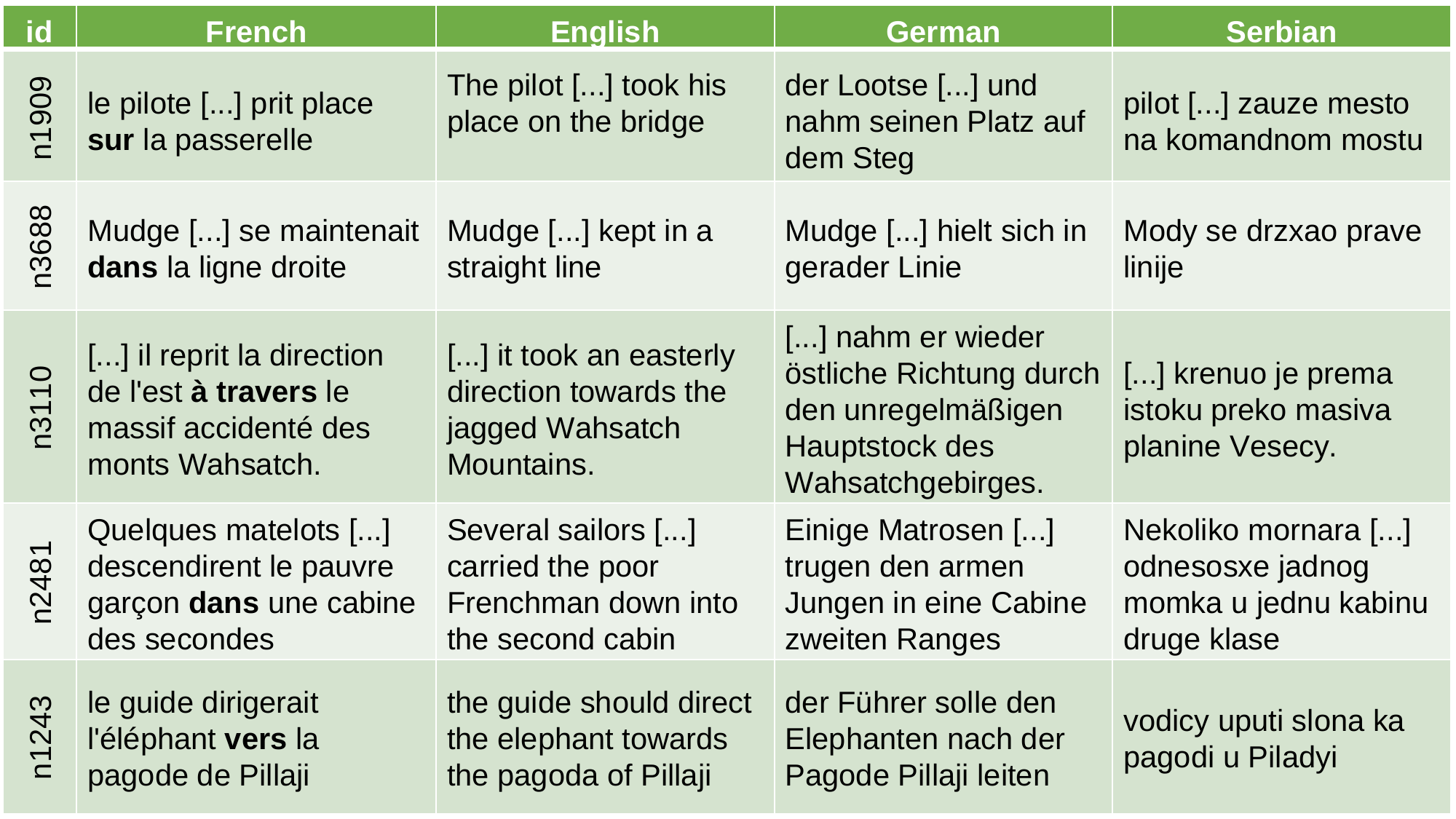}
\captionsetup{type=table}
\caption{Excerpt of \citep{corpus}. Spatial prepositions in the French source sentences are emphasised. \textsf{id} refers to the original identifier as released in the corpus.}
\label{tab:corpus}
\end{center}
\end{figure*}

We propose to regard a word in context as a stimulus in its own right and to perform an IB analysis in the semantic domain of spatial relations. To do so, we use a novel type of data: bitexts, e.g. translations of source sentences into target ones (illustrated in Table~\ref{tab:corpus}). We frame the act of translation as a constrained optimisation problem in the IB sense, and assume that human translators aim for exact meaning preservation. A word of interest from a source sentence is then considered a stimulus, which triggers its own mental representation, which is then linguistically encoded in the target language. 

We tested this new methodology on a cornerstone of spatial language: spatial prepositions. These terms overtly mark for spatial relations between entities, and are highly prone to engage spatial representations \citep{bbs}. We first extracted cross-linguistic variations of these spatial terms in four languages, namely detecting occurrences in source sentences from \citet{corpus}, and aligning them with their target counterparts. This allowed us to assess the complexity of spatial prepositions translations. Furthermore, to estimate the informativity of these translations, we also modelled the psychological space of spatial relations by leveraging pile-sorting results from a pilot study ($N=35$). We show that a low-dimensional space suffices to predict similarity judgements. Ultimately, we provide preliminary evidence for greater communicative efficiency of natural translations compared to random and counterfactual systems, in the semantic domain of spatial relations.

\section{Theoretical framework}

This section outlines how the IB theory has been applied to language. Tracing back to the Rate-Distortion theory \citep{infotheo}, communication between a speaker and a listener is formalised as a joint search for an encoder and a decoder that enable lossy data compression. In this schematic communicative scenario, the speaker seeks a linguistic encoder $q$ that maps meanings to forms, which is minimally complex, and simultaneously minimises the \textit{distortion} between the intended meaning $M$, and the listener's reconstructed meaning $\hat{M}$.
An operationalisation of this trade-off was first proposed by \citet{tishby2000}, and later applied to colour-naming \citep{colours}, and other semantic domains \citep{mollica2021forms, demonstratives, gaits}.

In this line of work, the Complexity of the stochastic encoder $q$ is quantified by the mutual information\footnote{\label{note:def}
The mutual information is a general measure of how related two random variables are. It measures the reduction in uncertainty in one variable when observing the other: $I(X;Y) = H(X) - H(X|Y) = H(Y) - H(Y|X)$, with $H(X)$ being the entropy of $X$. Note that the joint distribution $p(X,Y)$ is all we need to compute the mutual information: \mbox{$I(X;Y) = \sum_{x,y} p(x,y) \log \frac{p(x,y)}{p(x)\cdot p(y)}$}. It is a symmetric measure: $I(X;Y)=I(Y;X)$.
}
$I_q(M;W)$ between two random variables: $M$ and $W$\footnote{We refer to random variables using upper-case letters, and realisations with lower-case letters.}. 
The mental representations the speaker wishes to communicate are denoted by $M$, and $W$ represents the discrete linguistic forms they are mapped onto, according to the policy $p_q(w|m)$ of the encoder $q$.
Conversely, the \textit{distortion} between the speaker's intended meaning $M$, and the listener's reconstructed meaning $\hat{M}$, is measured using the expected Kullback-Leibler distance,
\mbox{$\mathbb{E}_q[ D(M || \hat{M}]$}. 
Minimising this distortion is mathematically equivalent to maximising the quantity $I_q(W;U)$, which is defined as the Accuracy of the encoder \citep{thesis}.
In this formulation, $U$ represents the possible states of the world that the speaker perceives and describes,  and about which the listener forms probabilistic inferences.

The core trade-off between the compression of mental representations (measured by Complexity) and the preservation of meaningful information (measured by Accuracy) translates directly to optimisation. For any fixed amount of Accuracy, the encoder and decoder are optimised to ensure that Complexity is minimised. Introducing a Lagrange multiplier $\beta$, this trade-off can be framed as a constrained optimisation problem with the following objective function: 
\begin{equation}
    \label{eq:objective}
    \mathcal{F}_\beta[q] = I_q(M;W) - \beta I_q(W;U)
\end{equation}
where $\beta$ controls the trade-off. Any incremental 1-bit growth in Accuracy of a given optimal encoder incurs a cost of $\beta$ bits in Complexity. All natural or artificial encoders can be located in the Accuracy-Complexity information plane. From the Data Processing Inequality it follows that Complexity serves as an upper bound for Accuracy. Consequently, any encoders lying above the optimal frontier defined by the set $\{\mathcal{F^*_\beta} \mid 0 \leq \beta < +\infty\}$ are unachievable under the assumptions of the IB framework.

\section{General methodology}
Within a specific semantic domain, an IB analysis relies on two types of data to compute the Complexity and Accuracy of an encoder $q$: naming data and human similarity judgements. Examples of naming data include colour names (when describing colour chips in \citet{colours}), verbs of human gait (when describing a person moving in short videos, in \citet{gaits}), object names (when describing real-world images in \citet{gualdoni2023}), or spatial relations (when describing drawings of spatial scenes in \citet{khetarpal}). In this work, we apply the IB framework to the translation task.
In this section, we detail how naming data and similarity judgements can be recovered from translation data, and how Complexity and Accuracy measures are derived.

Using a bitext, we treat each occurrence of a term of interest within its sentential context $U$ as a stimulus that evokes a meaning representation $M$. Consider, for example, $u = \textsf{the apple \underline{in} the bowl}$. Here, the spatial relation between the apple and the bowl is not only determined by the lexical item $\textsf{\underline{in}}$ alone, but emerges from the full sentential context in which it is embedded.
Crucially, two occurrences of the same lexical item evoke the same meaning $M$ if and only if they appear in identical sentential contexts, that is, when the same phrase is repeated in the bitext corpus. Under this assumption, $M$ is contextually grounded rather than lexically fixed. The evoked representation $M$ is then encoded using a lexical item $W$ in the target language. The schema \mbox{$U \rightarrow M \rightarrow W$} summarises how translations are generated under these assumptions.

\subsection{Naming data}
\begin{figure}[!ht]
\begin{center}
\includegraphics[width=\columnwidth]{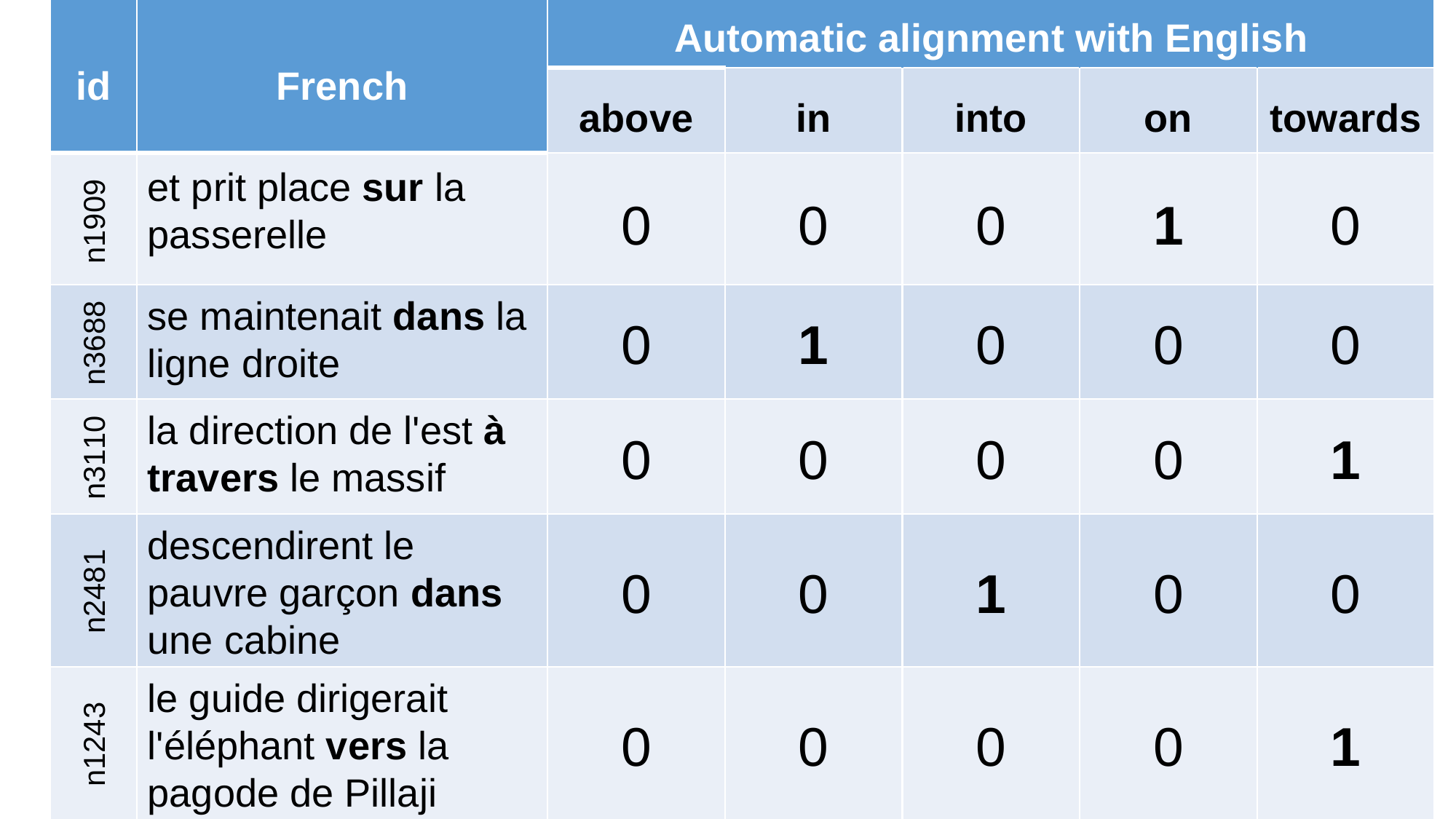}
\captionsetup{type=table}
\caption{French to English excerpt of computed alignments of spatial prepositions from \citet{corpus}. Each row is a one-hot encoding of the source preposition over the target lexicon. The source preposition is emphasised in bold.}
\label{tab:alignement}
\end{center}
\end{figure}

Given a bitext corpus, our goal is to align terms of interest from source sentences with their counterparts from target sentences. For example, the French spatial prepositions from Table~\ref{tab:corpus} ($u_1 = \textsf{le pilote prit place \underline{sur} la passerelle}$ and $u_2 = \textsf{Mudge se maintenait \underline{dans} la ligne droite}$) map to their English equivalents $w_1=\textsf{on}$ and $w_2=\textsf{in}$.
This process yields alignment tables like the one presented in Table~\ref{tab:alignement}. 
From these alignments, we can derive the conditional probability distribution $p_q(w|m)$ where $m$ ranges over each distinct meaning and $w$ ranges over the target lexicon. Assuming a uniform prior distribution $p(m)$ over all meanings in the source text
\footnote{\citet{colours} show that replacing an empirically derived prior reflecting actual communicative need with a simple uniform prior has little qualitative influence on the core finding. We use therefore a uniform prior distribution in this first step 
}, we compute the joint distribution: \mbox{$p_q(w,m) = p_q(w|m)p(m)$}.
From this, we derive the Complexity $I_q(M;W)$ (see footnote 1 for details).

\subsection{Similarity judgements}

Collecting contextual similarity judgements of source terms grows quadratically in the size of the bitext. Being able to estimate the pairwise similarity between any pair of source terms is therefore crucial. As part of our proposed methodology, we first
select a subset of terms in their sentential context \mbox{$\{ u_i \mid 1\le i \le K\}$} of reasonable size $K$, and collect human pairwise similarity judgements \mbox{$\{ sim(u_i,u_j) \mid 1 \le i,j\le K \}$}. For ease of exposition, we denote these empirical similarities as $sim_{i,j}$. Based on these pairwise similarity judgements, we then
train a model to predict the similarity $\widehat{sim}_{i,j}$ 
for any new unseen pair of terms.
Following \citet{colours} and \citet{demonstratives}, we derive a subjective belief distribution on world states conditional on meanings $p(u|m)$, which is estimated as 
\begin{equation}
    p(u_j|m_i) \propto \exp\left(\gamma \cdot \widehat{sim}_{i,j}\right)
\end{equation}
Here, $\gamma$ is a temperature parameter that controls the penalty for mistaking two distinct stimuli\footnote{We set $\gamma$ so that $p(u|m)$ is defined by a softmax over similarity scores. Alternative methods for grounding the distribution $p(u|m)$ should be explored in future work.}.
Note that when IB models a subjective belief as a distribution on world states $U$, it captures the ambiguity a term of interest can have in its context.
Using the distributions $p(u|m)$, $p_q(w|m)$, and $p(m)$, we can now compute the joint distribution: \mbox{$p_q(w,u) = p_q(w|m)p(m)p(u|m)$}. From this, we derive the Accuracy $I_q(W;U)$.

\section{Experiments}

For the remainder of this work, we focus on the specific semantic domain of spatial relations. These relations are typically marked by prepositions as discussed extensively in the literature (for French linguistic aspects we refer to \citet{vandeloise, borillo, lepesant}, for a broad discussion on spatial cognition we refer to \citet{bbs}). 
We also consider \say{absolute use} of prepositions \citep{lepesant}, which refers to instances where a preposition appears without its standard complement because the referent is understood from context (e.g., $\textsf{Je me gare devant}$, $\textsf{I park in front}$, $\textsf{Ich parke davor}$ and $\textsf{Parkiram ispred}$). More importantly, we consider a preposition to have a spatial sense when its surrounding context triggers a concrete spatial mental imagery, involving entities that can be physically touched or pointed at. 
This operational definition is theoretically grounded in the notions of \textit{cline of concreteness} \citep{concreteness} and \textit{cline of collocation} \citep{mip}.

Throughout our experiments\footnote{Data and code are available at \url{ https://github.com/antoineOYO/spatial-terms-translations}}, we utilised English, German, and Serbian translations of Jules Verne's French adventure novel, \textit{Le tour du monde en 80 jours}. The sentence-level alignment across these four texts was previously processed by \citet{corpus}.

\subsection{Naming data: detection and alignment of spatial prepositions}\label{sec:align}

We first collected occurrences of prepositions with potential spatial sense, namely searching for instances from the lexical inventory of spatial prepositions established by \citet{lepesant}. An initial disambiguation of each occurrence was then carried out jointly by three annotators until consensus was reached. The lexical and semantic disambiguation steps yielded 1,312 prepositions with a spatial sense. Using the Large Language Model \texttt{mistral-large-2512} \citep{mistral} in a few-shot setting, these French prepositions were subsequently aligned with their English, German and Serbian counterparts. The alignment table for Table~\ref{tab:corpus} is illustrated in Table~\ref{tab:alignement}. 
The few-shot prompt for French-to-English alignment is available in Appendix~\ref{app:fewshot}. 
To evaluate the quality of these automatic alignments, we sampled 100 preposition pairs for each target language. Errors of alignment were defined as the alignments of source prepositions to either inadequate target prepositions, or to any distinct other parts of speech, such as nouns or verbs. 
This evaluation yielded precision scores of 81\% for French–English, 91\% for French–German, and 94\% for French–Serbian. Out of the 1,312 source prepositions, 580 were successfully aligned across all three target languages. The remaining instances were discarded, primarily because the source preposition was not relexified with a preposition in at least one of the translations\footnote{This is also known as \textit{zero-marking}: consider the use of the preposition in $\textsf{Je suis à la maison}$ versus $\textsf{I'm home}$. We refer to \citet{bible} for related analysis.}. The cross-linguistic discrepancy between automatic alignments is discussed in section~\ref{sec:discopti}.

Furthermore, we computed contextual embeddings for each source preposition, using \texttt{xlm-roberta-large} \citep{xlm}. We averaged and stacked the last four hidden layers, yielding embeddings of size $d=4,096$. We then applied $K$-means clustering ($K=30$) to obtain a representative subset of 30 French sentences, each with a spatial preposition, for downstream analysis. We found that $K=30$ reflected both the distribution of French spatial terms in the corpus, and the embedding space
(see Appendix~\ref{app:tsne} for a dimension reduction). The clustering and dimension reduction were implemented using \texttt{scikit-learn} \citep{scikit-learn}.

\input{triple-figure.tex}

\subsection{Models of human similarity judgements}
Our objective is to predict the human-perceived similarity between two distinct spatial relations expressed by spatial prepositions from any source sentence.

For this purpose, we designed a pile-sorting task based on the $K=30$ samples obtained from our clustering step.
To assess the feasibility of these similarity judgements, we conducted an anonymised pilot study (see section \ref{ethics}) with a cohort of colleagues ($N = 35$), all of whom were native French speakers. The reported patterns should be interpreted as preliminary evidence. The 30 samples were printed on individual cardboard cards. The participants were orally instructed to group the samples into as many piles as they deemed necessary based on their perceived similarity; there was no time limit and participants could revise their groupings until satisfied. The full instructions (see Appendix~\ref{app:instructions}) were printed and available to each participants throughout the whole task. The distribution of number of piles produced by the participants is displayed in Appendix~\ref{app:distrib_piles}.

Following \citet{khetarpal}, we define the pairwise similarity $sim_{i,j}$ as the proportion of all participants who sorted $u_i$ and $u_j$ into the same pile. A visualisation of the proportions is shown in Figure~\ref{fig:matrix}. From the list of pairwise similarities, a Multidimensional Scaling (MDS)\footnote{Note that this MDS is for visualisation purpose only.} was then computed using \texttt{scikit-learn} implementation (Figure \ref{fig:mds}). Both visualisations suggest that the similarity structure of spatial relations is complex and graded rather than discrete or reducible to surface form (discussed in section~\ref{ref:discsim}). 

We modelled these similarity judgements using three approaches based on the contextual embeddings $F_i$ of size $d$ we computed in the previous step. As a baseline, we computed the Cosine Similarity between z-scored embeddings, following \citet{zscore}. We compared this baseline against two learned models: a Ridge regression and a low-rank projection model (replicating \citet{lowdim}). The Ridge regression model is a simple linear regression, where the size of the coefficients is penalised with a parameter $\alpha$. It is trained on the vertical stack of three features: the Hadamard product $F_i \odot F_j$, their L1 distance, and their cosine similarity. Conversely, the low-rank projection model learns a mapping matrix $P \in \mathbb{R}^{D \times d}$ to project embeddings into a $D$-dimensional space, predicting similarity as $\widehat{sim}_{i,j} = (P F_i)^\top (P F_j)$. This model minimises the mean squared error loss with L2 regularisation:
\begin{equation}
    \min_{\lambda,P}\; \frac{1}{2}\sum_{i,j}||sim_{i,j} - \widehat{sim}_{i,j}||^2 + \lambda||P||^2
\end{equation}

Hyperparameters (including the Ridge $\alpha$, the projection rank $D$, and the regularisation penalty $\lambda$) were optimised within 6-fold inner validation splits, and the generalisation error (measured via Spearman's $\rho$) was computed across 6 outer test folds using a nested cross-validation procedure.
Model performance (see Table~\ref{tab:regression}) is reported using Spearman's rank correlation coefficient $\rho$, as it is better suited for evaluating monotonic relationships than Pearson's $r_P$, which evaluates strictly linear ones \citep{spearman}. Results presented in Table~\ref{tab:regression} indicate that a simple cosine similarity does not strongly correlate with the sortings we collected, whereas the low-rank projection model achieves a high Spearman correlation of $\rho = 0.7800 \pm 0.0622$, with dimension $D=5$.
\input{regression.tex}

\begin{figure*}[!ht]
\begin{center}
\begin{subfigure}[t]{0.49\textwidth}
\centering
\includegraphics[width=\textwidth]{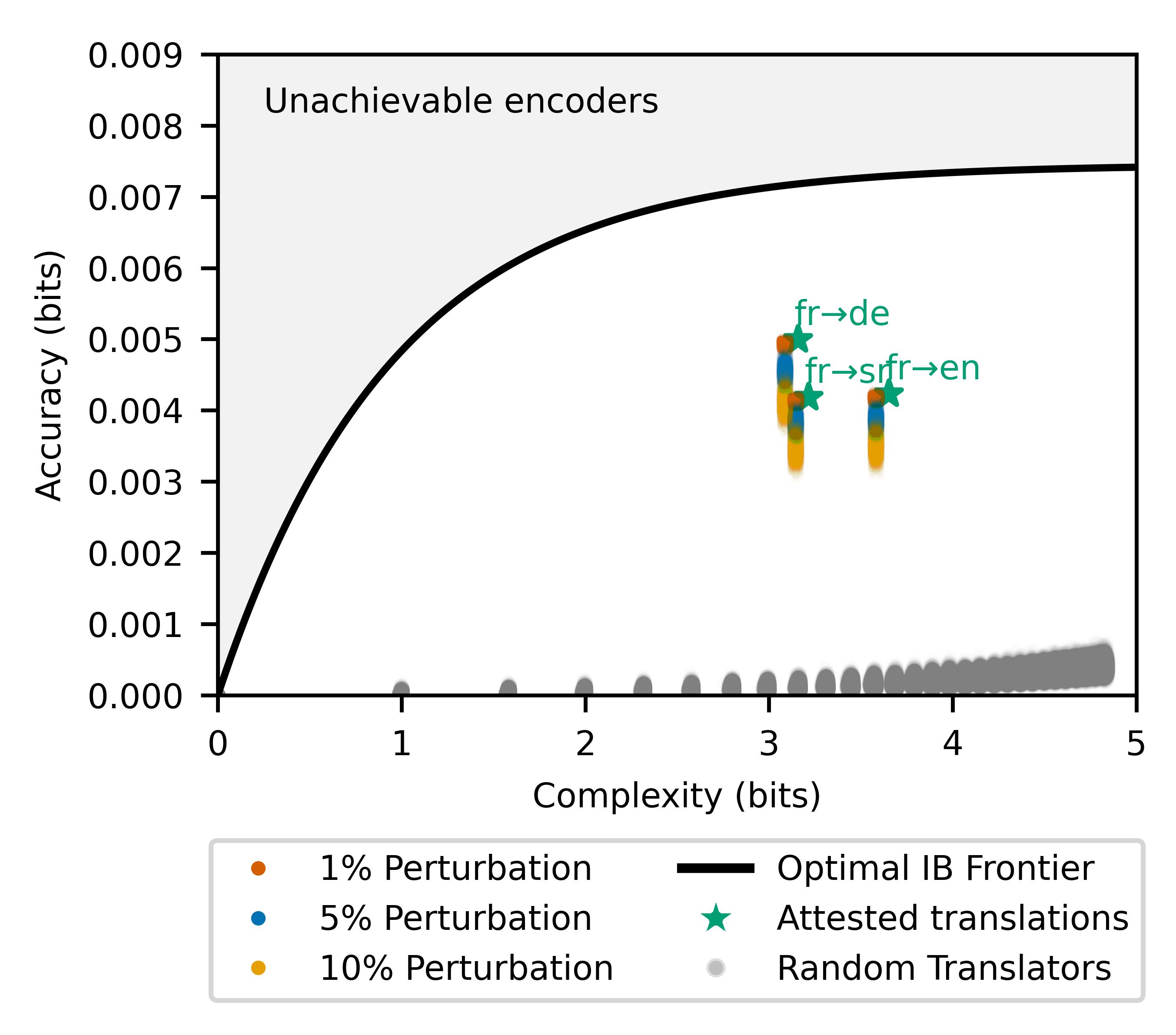}
\caption{Information plane of attested, perturbed and random encoders of spatial prepositions. Dots are jittered slightly along the Complexity axis for visualisation purposes.}
\label{fig:infoplane}
\end{subfigure}
\hfill
\begin{subfigure}[t]{0.49\textwidth}
\centering
\includegraphics[width=\textwidth]{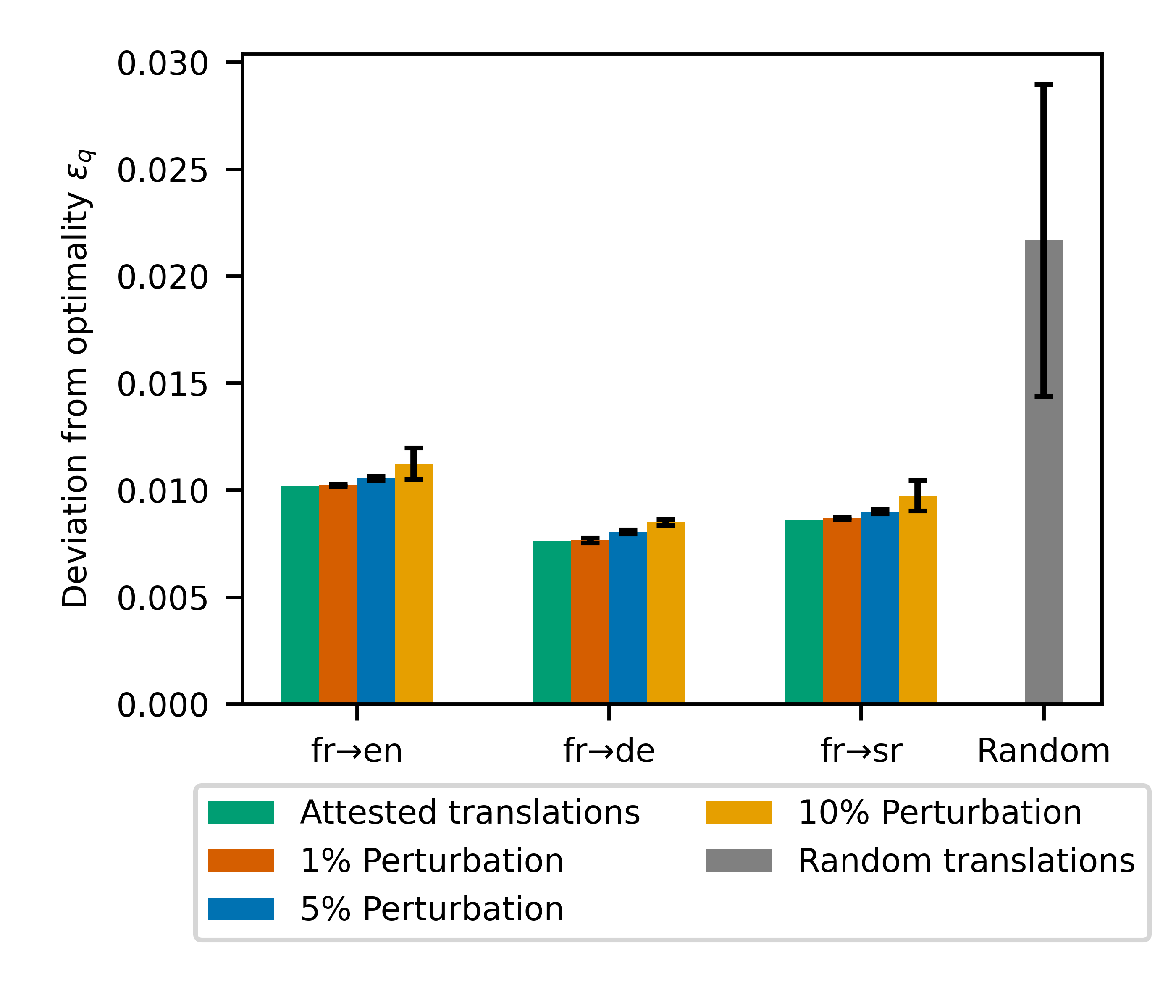}
\caption{Deviation from optimality for attested, perturbed and random encoders. Black error bars indicate standard deviations.}
\label{fig:deviations}
\end{subfigure}
\caption{How much do attested and random systems deviate from optimality?}
\label{fig:combined}
\end{center}
\end{figure*}

\subsection{Computation of encoders locations in the information plane}

To benchmark the attested translation encoders, we generated 3x10,000 counterfactual \say{perturbed encoders} by randomly permuting 1\%, 5\% and 10\% of the rows within the alignment tables. We also generated 300,000 uniform-random translation encoders. We then computed the Accuracy and Complexity for attested, perturbed and fully random translation encoders using the implementation provided by \citet{colours}. The optimal frontier was computed using the reversed deterministic annealing algorithm implementation from \citet{demonstratives}, and was traced by iterating over 100 values of $\beta$ log-linearly spaced between 1 and $2^{20}$. Attested, perturbed and random translations, along with the optimal frontier, are displayed in the information plane from Figure~\ref{fig:infoplane}. We observe that attested translation encoders lie far from the steepest part of the theoretical bound, which is discussed in section \ref{sec:discopti}.

Within the information plane, the distance of any given encoder $q$ to the optimal frontier defines its deviation from optimality, as established by \citet{colours}:
\begin{equation}
\epsilon_q = \min_\beta \frac{1}{\beta}\left(\mathcal{F}_\beta[q] - \mathcal{F}^*_\beta \right),
\end{equation}
The deviations of translation encoders is shown in Figure~\ref{fig:deviations}. Results indicate that as the degree of perturbation increases, the efficiency of translation encoders decreases.

\section{Discussion} \label{sec:discussion}

\subsection{Characterising cross-linguistic use of spatial prepositions}
The automatic detection and alignment of spatial prepositions in bitexts has received limited attention until recent studies such as those by \citet{bitexts} and \citet{bible}. For detection, we handled the bitext differently by choosing not to apply a syntactic filter to the source sentences. This choice is based on the observation that spatial terms appear in a variety of syntactic contexts, and range from single terms ($\textsf{sur}$, $\textsf{dans}$), to nominal locutions ($\textsf{à bord du}$, $\textsf{au milieu de}$). With respect to alignment, \citet{bitexts} rely on the \textsf{GIZA++} word alignment tool \citep{gizza}. In contrast, the RUIMTE model proposed by \citet{bible} avoids traditional word alignment, instead identifying spatial relation markers by calculating statistical associations. Direct performance comparisons between these methods are difficult, as these previous works evaluated highly morphologically complex languages. While our pipeline introduces a dependency on few-shot extraction tools, it generalises to any bitext without requiring morphologically complex language-specific alignment models.

\subsection{Estimation of similarity between spatial relations evoked by prepositions}\label{ref:discsim}

The correlation between judgements of word sense relatedness and computational representations relatedness has been widely investigated using both static embeddings \citep{hill2015}, contextual embeddings and Large Language Models \citep{nair2020, marjieh2022words}. Specifically, how context modulates the perceived meaning similarity between word pairs in written sentences has been graded by human subjects in many experiments \citep{huang-etal-2012-improving, pilehvar-camacho-collados-2019-wic, armendariz-etal-2020-cosimlex, vulic2020multi}.
However, most of these studies focus on nouns, adjectives or verbs. Our pilot study, conversely, was concerned with relations between entities, and the prepositions that mark them. 

Results presented in Table~\ref{tab:regression} indicate that a simple cosine similarity does not strongly correlate with our collected human sortings, whereas the low-rank projection model achieves a high Spearman correlation of $\rho = 0.7800 \pm 0.0622$. Interestingly, low-rank projection models achieve the highest correlation scores with $D=5$. This suggests that higher-dimensional projections likely lead to overfitting. These findings indicate that low-rank projections may be more effective than standard cosine similarity for predicting contextual similarities in spatial relations.

Aside from modelling, we acknowledge that our subset of 30 spatial prepositions in context does not guarantee cognitive representativeness of the domain of spatial relations. However, results from Figure~\ref{fig:matrix}~and~\ref{fig:mds} suggest that colexification does not explain alone the similarity structure of spatial relations. We assume that modelling the semantic similarity of spatial relation in a continuous $D$-dimensional space accounts for the graded similarity space we observed, and that a sufficiently large bitext should provide broad coverage of the domain.

However, a lexical bias may arise during the pile sorting task, where shared word forms inflate human judgements of semantic similarity. As a result, similarity ratings may partly reflect surface-form overlap rather than purely relational semantics. This introduces a potential language-specific skew, whereby the lexical structure of French influences perceived relatedness. To mitigate this effect, future work should control for form overlap in the stimuli and compare across source languages, in order to quantify the contribution of lexical structure to similarity judgements.

We also acknowledge that this subset is not fully representative of all linguistic means available for encoding spatial relations. While we focused on prepositions as a first step, alternative encoding mechanisms must be accounted for in future research, such as interactions between the preposition and the verb (see for example the verb-framing typologies from \citet{Talmy1983} and \citet{aurnague2012espace}).

\subsection{Efficiency of attested systems} \label{sec:discopti}
In Figure~\ref{fig:infoplane} we observe that attested translations of spatial prepositions lie closer to optimal systems than random ones. To further investigate whether this is a feature of \say{reasonable} translations, we generated counterfactual translations with varying degrees of perturbation. The results from Figure~\ref{fig:deviations} indicate that as the degree of perturbation increases, the deviation to optimality increases. We believe that this comparison would benefit from including another baseline, such as a surjective mapping from meanings to terms, as used by \citet{dictionary}. This is equivalent to removing the non-deterministic assumption on the encoder $q$ \citep{strouse2017deterministic}. Future work should also test how altering the similarity geometry affects Accuracy. Nevertheless, these results suggest that attested systems exhibit a pressure for efficiency under the IB framework.

We conclude by addressing why attested encoders deviate from optimality, and the confounds inherent to translation data. This question lies at the crossroads of research on language efficiency and translation process \citep{teich_translation_2020}. Translating involves genre conventions and notably stylistic incentives. This is illustrated by observations from the corpus: while German and Serbian translations of \textit{Le Tour du monde en 80 jours} remain stylistically conservative, the English version acts as a \say{free} translation with added details. This divergence is clear in the original corpus, where \citet{corpus} found and explicitly annotated changes in meaning, but were found only in the English version. From an information-theoretic perspective, these changes in meaning represent a deviation from strict communicative efficiency, as the translator prioritised semantic elaboration or expressivity over the compact encoding predicted by the IB optimum. Yet, a conservative translator might also imitate some of the lexical choices of the French original despite what would be natural in their language \citep{volansky2015features, lim_simpsons_2024}. The alignment of preposition-to-preposition we performed (see section~\ref{sec:align}) is by design too coarse to capture the aforementioned phenomena. Therefore we argue that IB analysis of translation data must narrow the scope to relexification of terms of interest, such as prepositions, in order to capture only one aspect of the translation process.

\section{Conclusion}

This work proposes a novel methodology for analysing communicative efficiency via translation data. We performed an IB analysis on spatial prepositions across three translations of a French novel.
Our preliminary results suggest that the translation of spatial prepositions exhibits pressure toward optimal trade-offs between informativity and complexity, consistent with the IB literature, and more broadly, recent evidence from language efficiency. An important next step is to apply this approach to typologically diverse, genealogically distant languages, allowing us to better separate general efficiency pressures from patterns arising through shared linguistic ancestry. We believe we presented a plausible way to extend to other semantic domains, other linguistic forms than prepositions, and scale beyond fully elicitepd naming datasets while staying grounded in natural speech.

\section{Ethics Statement} \label{ethics}
Written informed consent was obtained from all participants. Pile-sorting data consisted only of lists of sentences. These data were fully anonymised at the point of collection, such that no personal data were retained; the data from the pile-sorting experiment therefore fall outside the scope of the General Data Protection Regulation (GDPR, Regulation (EU) 2016/679). Given the minimal-risk nature of this feasibility study (involving no sensitive information and no identifiable data) institutional ethics committee approval was not sought.

\section{Acknowledgements}
We sincerely thank our colleagues at Laboratoire d'InfoRmatique en Image et Systèmes d'information for participating in the pile-sorting experiment. We further thank the participants of the SIGTYP workshop for their stimulating discussions and valuable feedback during the presentation of this work. We are also grateful to the anonymous reviewers for their constructive suggestions, which helped us improve this paper.

\section{Bibliographical References}\label{sec:reference}

\bibliographystyle{plainnat}
\bibliography{custom}

\appendix
\section*{Appendix}
\addcontentsline{toc}{section}{Appendix}

\section{Prompt for French to English few-shot alignment}\label{app:fewshot}

\input{prompt.tex}

%\clearpage
\section{Instructions for pile sorting}\label{app:instructions}
\input{instructions.tex}

%\clearpage
\section{Pairwise similarity modelling}\label{app:distrib_piles}

\renewcommand{\thefigure}{C-\arabic{figure}} 
\setcounter{figure}{0} 

\begin{figure}[H]
\begin{center}
\includegraphics[width=\columnwidth]{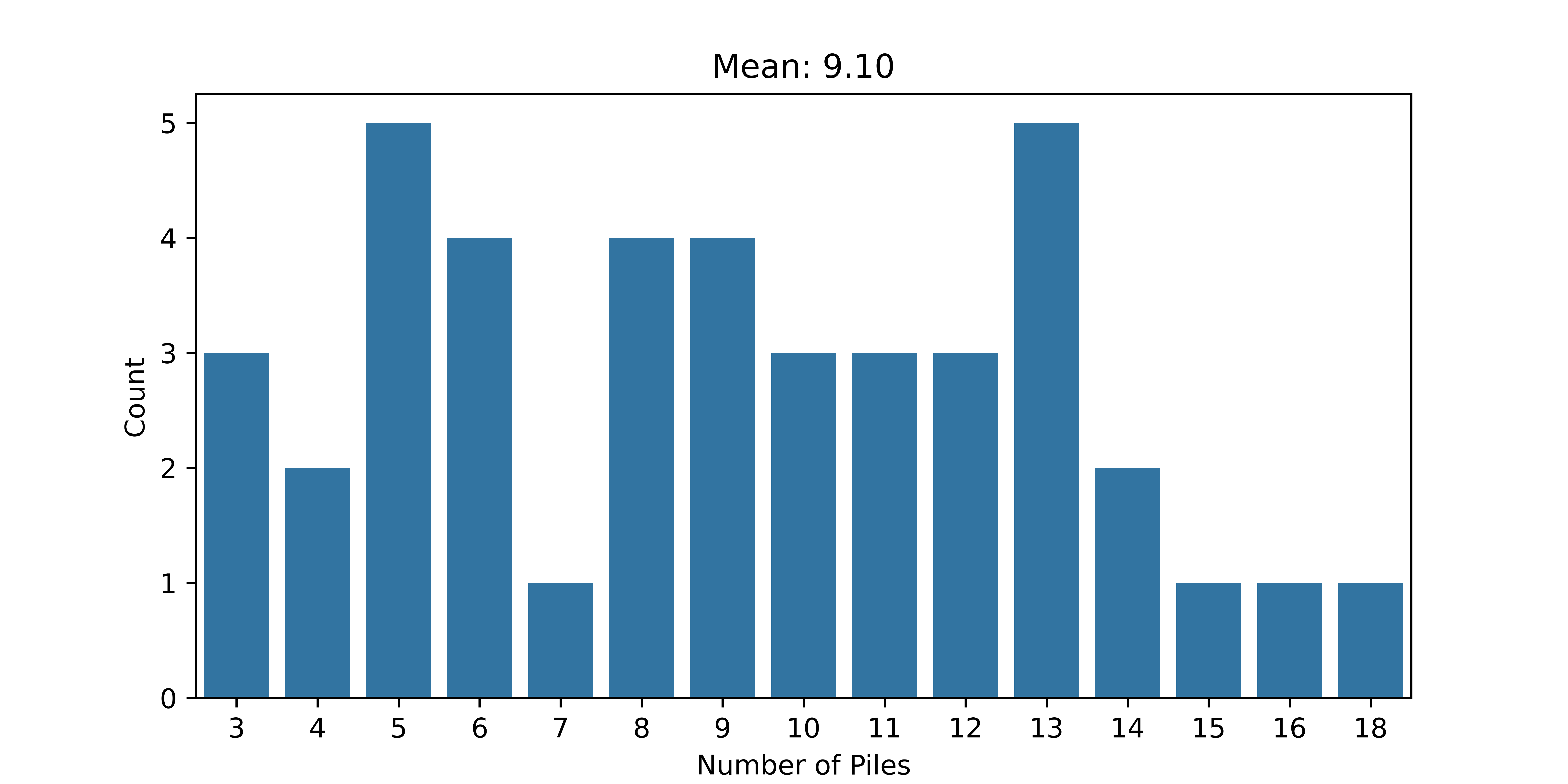}
\caption{Distribution of the number of piles across participants (N=35).}
\label{fig:distrib_piles}
\end{center}
\end{figure}

\clearpage
\onecolumn
\section{Dimension reduction of the embedding space}\label{app:tsne}

\renewcommand{\thefigure}{D-\arabic{figure}} 
\setcounter{figure}{0}

\begin{figure*}[!ht]
\centering
\includegraphics[width=\textwidth, trim= 0 5cm 0 1cm, clip]{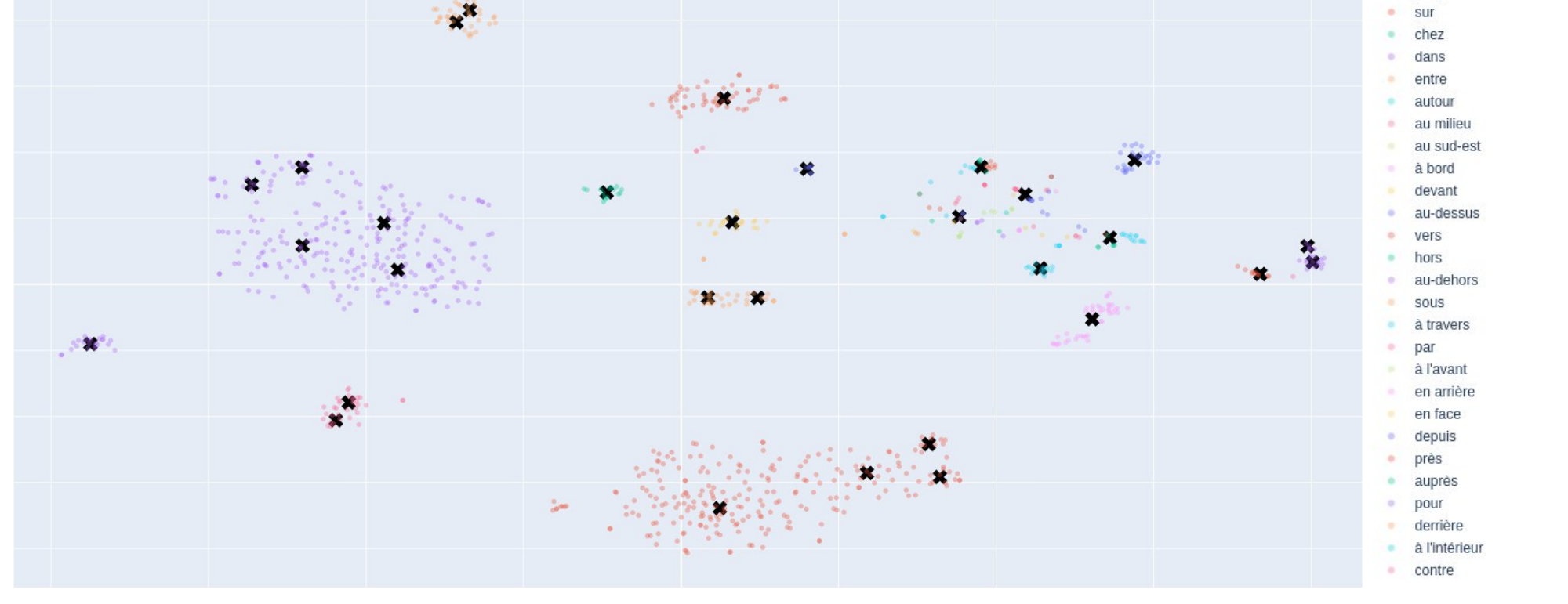}
\caption{t-SNE visualisation of the 30
K-means centroids (computed from stacked final-layer representations of \texttt{xlm-roberta-large}) and embeddings of the most frequent French prepositions}
\label{fig:tsne}
\end{figure*}

\end{document}

%% file: triple-figure.tex
 \begin{figure*}[t] 
      \centering
      \begin{subfigure}[t]{0.48\textwidth}
          \centering
          \includegraphics[width=0.83\textwidth]{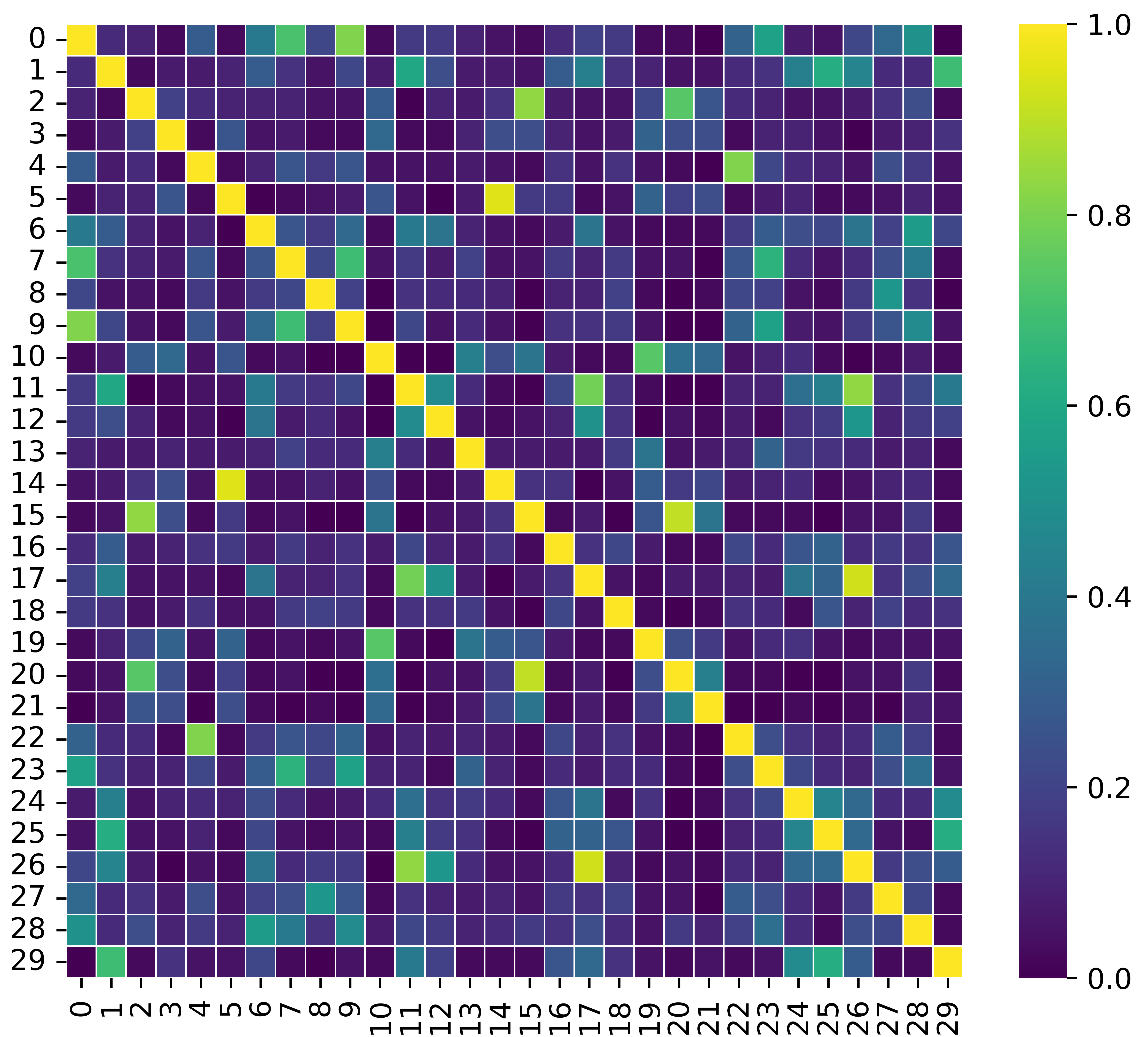}
          \caption{\label{fig:matrix}Proportion of participants (N=35) who grouped 2 spatial relations into the same pile.}
      \end{subfigure}
      \hfill
      \begin{subfigure}[t]{0.48\textwidth}
          \centering
          \includegraphics[width=\textwidth]{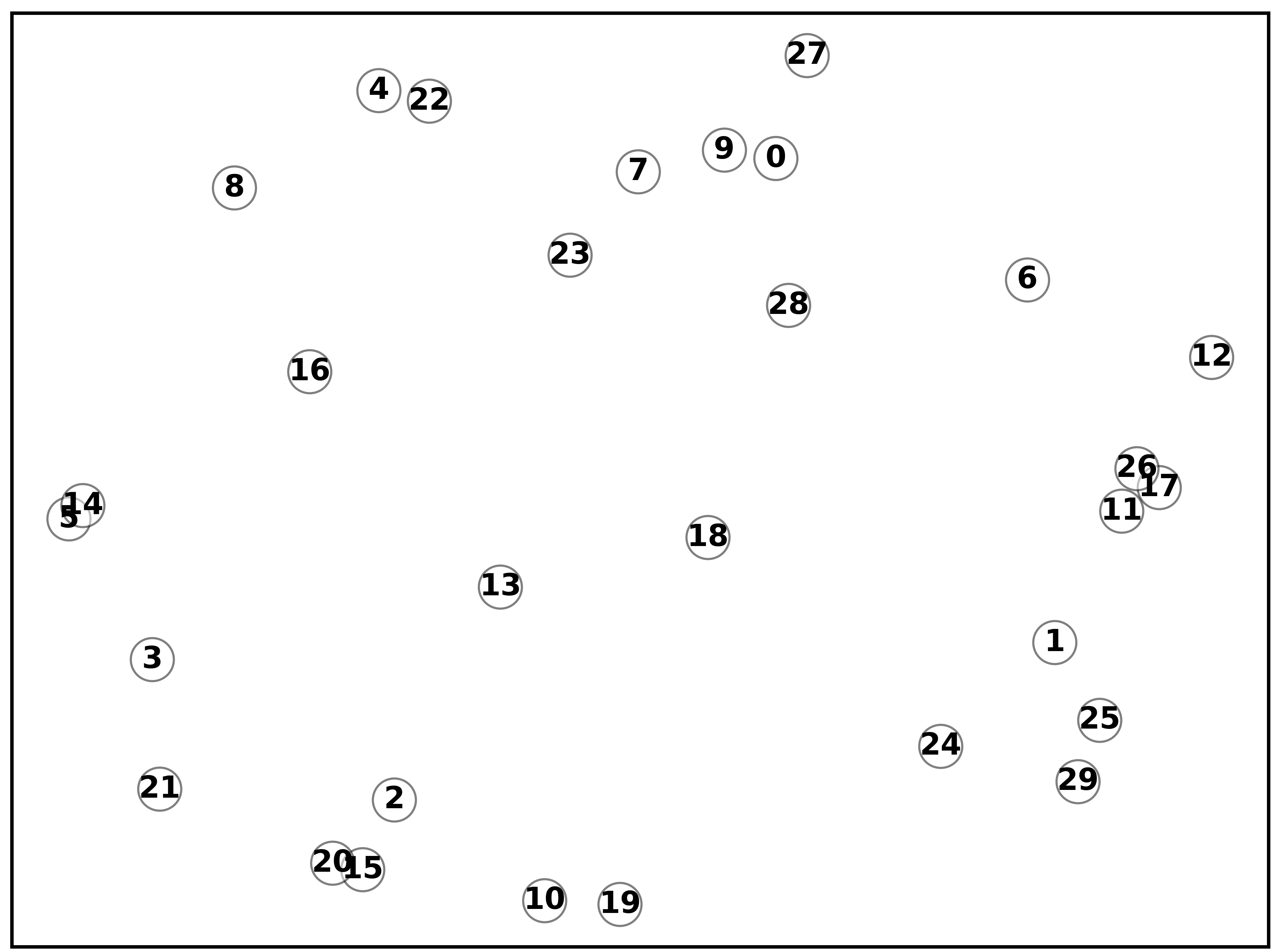}
          \caption{\label{fig:mds}MDS visualisation of the pairwise similarities.}
      \end{subfigure}
      \vspace{1.5em} % Espacement vertical
      \begin{subfigure}[b]{0.98\textwidth}
          \centering
          \includegraphics[width=.95\textwidth]{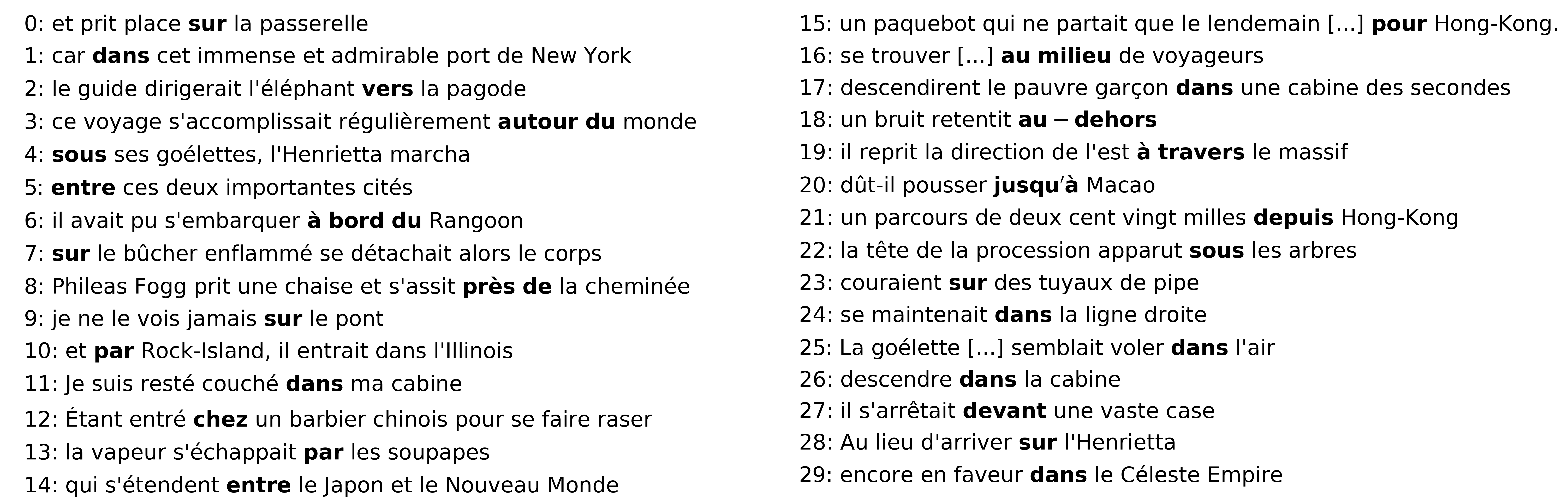}
          \caption{\label{fig:index}Prepositions (emphasised) and context of the 30 spatial relations sorted by the participants}
      \end{subfigure}
      \caption{Empirical similarity data from the pile-sorting task.}
      \label{fig:pile_sorting}
 \end{figure*}

%% file: regression.tex
\begin{table}[ht]
    \centering
    \begin{tabular}{lc}
        \toprule
        \textbf{Model} & \textbf{Spearman $\rho$} \\
        \midrule
        Cosine Similarity (Baseline) & $0.079$ \\
        Ridge Regression & $ 0.647 \pm 0.056$ \\
        % \midrule
        \multicolumn{2}{l}{Low-rank Projection} \\
        \hspace{3mm} $D=1$ & $ 0.263 \pm 0.140$ \\
        \hspace{3mm} $D=5$ & $ \textbf{0.780} \pm \textbf{0.062}$ \\
        \hspace{3mm} $D=10$ & $ 0.762 \pm 0.053$ \\
        \hspace{3mm} $D=15$ & $ 0.751 \pm 0.053$ \\
        \hspace{3mm} $D=20$ & $ 0.749 \pm 0.034$ \\
        \hspace{3mm} $D=50$ & $ 0.648 \pm 0.070$ \\
        \hspace{3mm} $D=100$ & $ 0.578 \pm 0.101$ \\
        \bottomrule
    \end{tabular}
    \caption{Spearman correlation between model-predicted pairwise similarities and human pile-sorting judgements. Models all take pairs of contextual embeddings as input. Learned models performance is computed from 6-fold nested cross validation.}
    \label{tab:regression}
\end{table}
%0.576757  0.101241

%% file: prompt.tex
{\raggedright
\# ALIGNMENT OF SPATIAL TERMS

\noindent You are an expert linguist and translator specialised in French and English.
French is always the source language, and the target language is English.
You will be provided with a source sentence in French, and the target sentence in English.
In the source sentence, a spatial term is wrapped with <loc> and </loc>.
This spatial term marks a concrete spatial relationship.

\noindent Your task: align the spatial term in the source sentence with its translation in the target sentence, if any.

\noindent How: wrap with <loc> and </loc> the corresponding spatial term in the target sentence, if any. Also return the normalized form of the target spatial term (e.g., "on the right of" to "on the right").

\noindent **Constraints:**

\noindent 1. Spatial terms only include prepositions, adverbs, and locutions.

\noindent 2. Zero-alignment: if you cannot find an adequate spatial term in the target sentence corresponding to the translation of the source spatial term, don't wrap anything in <loc> from the target sentence and return "None" for the normalized form.

\noindent 3. If the meaning of the source spatial term is translated only by a verb or a noun in the target sentence, don't wrap anything in <loc> from the target sentence and return "None" for the normalized form. 

\noindent \#\# Example 1: basic alignment

\noindent INPUT = \{
    "source\_sentence" : "Passepartout demeura seul <loc>dans</loc> la maison de Saville-row",
    "target\_sentence": "Passepartout remained alone in the house of Saville-row",
\}

\noindent OUTPUT = \{
    "target\_sentence": "Passepartout remained alone <loc>in</loc> the house of Saville-row",
    "target\_normalized\_term": "in"
\}

\noindent \#\# Example 2: zero-alignment

\noindent INPUT = \{
    "source\_sentence" : "Il rentra <loc>chez</loc> lui à dix heures.",
    "target\_sentence": "He returned home at ten o'clock.",
\}

\noindent OUTPUT = \{
    "target\_sentence": "He returned home at ten o'clock.",
    "target\_normalized\_term": "None"
\}

\noindent \#\# Example 3: complex locution

\noindent INPUT = \{
    "source\_sentence" : "Au moment où l'Henrietta appareillait,tous quatre étaient <loc>à bord</loc>.",
    "target\_sentence": "all four were aboard at the moment when the Henrietta was setting sail.",
\}

\noindent OUTPUT = \{
    "target\_sentence": "all four were <loc>aboard</loc> at the moment when the Henrietta was setting sail.",
    "target\_normalized\_term": "aboard"
\}

\noindent \#\# Example 4: zero-alignment 

\noindent INPUT = \{
    "source\_sentence" : "Il a marché <loc>juqu'à</loc> la ville.",
    "target\_sentence": "He reached the city by foot.",
\}

\noindent OUTPUT = \{
    "target\_sentence": "He reached the city by foot.",
    "target\_normalized\_term": "None"
\}

\noindent \# Now is your turn. Provide the output in JSON format as in the examples.
\par}

%% file: instructions.tex
\textbf{French instructions (as printed for the participants):} "Sur chaque billet est exprimée une relation spatiale entre 2 entités. Une préposition en gras vous signale cette relation. Formez des piles regroupant les relations spatiales qui expriment des configurations spatiales similaires. Vous pouvez faire autant de piles que vous le souhaitez (y compris des piles d'un seul billet si une relation vous semble unique). Le numéro sur les billets est aléatoire."
\newline
\textbf{A possible English translation:} "On each card, a spatial relationship between 2 entities is expressed. A bolded preposition indicates this relationship. Form piles grouping the spatial relationships that express similar spatial configurations. You can create as many piles as you like (including stacks with a single card if a relationship seems unique). The numbers on the cards are random."